\documentclass[runningheads]{llncs}
\usepackage{graphicx}
\usepackage{comment}
\usepackage{amsmath,amssymb,bbm,stmaryrd} 
\usepackage{color}
\usepackage{pdfpages}

\usepackage{hyperref}
\hypersetup{
    colorlinks=true,    
    urlcolor=magenta,
}

\newcommand{\set}[2]{\{#1\ |\ #2\}}
\newcommand{\norm}[1]{\|#1\|}
\newcommand{\indicator}[1]{\mathbbm{1}\llbracket #1 \rrbracket}
\newcommand{\ceil}[1]{\left\lceil{#1}\right\rceil }
\newcommand{\reffig}[1]{Fig.\,\ref{fig:#1}}
\newcommand{\reftab}[1]{Table \ref{tab:#1}}
\newcommand{\refsec}[1]{Section \ref{sec:#1}}
\newcommand{\refeq}[1]{Eq.\,\ref{eq:#1}}
\DeclareMathOperator*{\argmax}{argmax}
\DeclareMathOperator*{\argmin}{argmin}

\begin{document}
\pagestyle{headings}
\mainmatter
\def\ECCVSubNumber{4625}  

\title{Iterative Distance-Aware Similarity Matrix Convolution with Mutual-Supervised Point Elimination for Efficient Point Cloud Registration} 

\titlerunning{Iterative Distance-Aware Similarity Matrix Convolution}
%
\author{Jiahao Li\inst{1} \and
Changhao Zhang\inst{2} \and
Ziyao Xu\inst{3} \and
Hangning Zhou\inst{3} \and
Chi Zhang\inst{3}}

%
\authorrunning{J. Li et al.}
%

\institute{Washington University in St. Louis, St. Louis, USA \\
\email{jiahao.li@wustl.edu} \and
Xi'an Jiaotong University, Xi'an, China \\
\email{cvchanghao@gmail.com} \and
Megvii Inc., Beijing, China\\
\email{\{xuziyao,zhouhangning,zhangchi\}@megvii.com}}


\maketitle

\begin{abstract}
In this paper, we propose a novel learning-based pipeline for partially overlapping 3D point cloud registration. The proposed model includes an iterative distance-aware similarity matrix convolution module to incorporate information from both the feature and Euclidean space into the pairwise point matching process. These convolution layers learn to match points based on joint information of the entire geometric features and Euclidean offset for each point pair, overcoming the disadvantage of matching by simply taking the inner product of feature vectors. Furthermore, a two-stage learnable point elimination technique is presented to improve computational efficiency and reduce false positive correspondence pairs. A novel mutual-supervision loss is proposed to train the model without extra annotations of keypoints. The pipeline can be easily integrated with both traditional (e.g. FPFH) and learning-based features. Experiments on partially overlapping and noisy point cloud registration show that our method outperforms the current state-of-the-art, while being more computationally efficient. Code is publicly available at \url{https://github.com/jiahaowork/idam}.
\end{abstract}

\section{Introduction}
\label{sec:introduction}

Point cloud registration is an important task in computer vision, which aims to find a rigid body transformation to align one 3D point cloud (source) to another (target). It has a variety of applications in computer vision, augmented reality and virtual reality, such as pose estimation and 3D reconstruction. The most widely used traditional registration method is Iterative Closest Point (ICP) \cite{icp}, which is only suitable for estimating small rigid transformation. However, in many real world applications, this assumption does not hold. The task of registering two point clouds with large rotation and translation is called global registration. Some global registration methods \cite{fgr,goicp} are proposed  to overcome the limitation of ICP, but are usually very slow compared to ICP. 

In recent years, deep learning models have dominated the field of computer vision \cite{deeplearning,alexnet,inception,resnet,resnet2}. Many computer vision tasks are proven to be solved better using data-driven methods based on neural networks. Recently, some learning-based neural network methods for point cloud registration are proposed \cite{pointnetlk,dcp,prnet}. They are capable of dealing with large rotation angles, and are typically much faster than traditional global registration methods. However, they still have major drawbacks. For example, DCP \cite{dcp} assumes that all the points in the source point cloud have correspondences in the target point cloud. Although promising, learning-based point cloud registration methods are far from perfect.

In this paper, we propose the \textbf{Iterative Distance-Aware Similarity Matrix Convolution Network (IDAM)}, a novel learnable pipeline for accurate and efficient point cloud registration. The intuition for IDAM is that while many registration methods use local geometric features for point matching, ICP uses the distance as the only criterion for matching. We argue that incorporating both geometric and distance features into the iterative matching process can resolve ambiguity and have better performance than using either of them. Moreover, point matching involves computing a similarity score, which is usually computed using the inner product or $L2$ distance between feature vectors. This simple matching method does not take into consideration the interaction of features of different point pairs. We propose to use a learned module to compute the similarity score based on the entire concatenated features of the two points of interest. These two intuition can be realized using a single learnable \textbf{similarity matrix convolution} module that accepts pairwise inputs in both the feature and Euclidean space.

Another major problem for global registration methods is efficiency. To reduce computational complexity, we propose a novel \textbf{two-stage point elimination} technique to keep a balance between performance and efficiency. The first point elimination step, \textbf{hard point elimination}, independently filters out the majority of individual points that are not likely to be matched with confidence. The second step, \textbf{hybrid point elimination}, eliminates correspondence pairs instead of individual points. It assigns low weights to those pairs that are probable to be false positives while solving the absolute orientation problem. We design a novel \textbf{mutual-supervision loss} to train these learned point elimination modules. This loss allows the model to be trained end-to-end without extra annotations of keypoints. This two-stage elimination process makes our method significantly faster than the current state-of-art global registration methods.

Our learned registration pipeline is compatible with both learning-based and traditional point cloud feature extraction methods. We show by experiments that our method performs well with both FPFH \cite{fpfh} and Graph Neural Network (GNN) \cite{dgcnn,rscnn,gnn} features. We compare our model to other point cloud registration methods, showing that the power of learning is not only restricted to feature extraction, but is also critical for the registration process.

\section{Related Work}
\label{sec:related-work}

\subsubsection{Local Registration}
\label{sec:local-related}
The most widely used traditional local registration method is Iterative Closest Point (ICP) \cite{icp}. It finds for each point in the source the closest neighbor in the target as the correspondence. Trimmed ICP \cite{trimmedicp} extends ICP to handle partially overlapping point clouds. Other methods \cite{efficienticp,generalizedicp,sparseicp} are mostly variants to the vanilla ICP.

\subsubsection{Global Registration}
\label{sec:global-related}
The most important non-learning global registration methods is RANSAC \cite{ransac}. Usually FPFH \cite{fpfh} or SHOT \cite{shot} feature extraction methods are used with RANSAC. However, RANSAC is very slow compared to ICP. Fast Global Registration (FGR) \cite{fgr} uses FPFH features and an alternating optimization technique to speed up global registration. Go-ICP \cite{goicp} adopts a brute-force branch-and-bound strategy to find the rigid transformation. There are also other methods \cite{convexrelaxation,integerprogramming,extremeoutlier,sdrsac} that utilize a variety of optimization techniques.

\subsubsection{Data-driven Registration}
\label{sec:data-related}
PointNetLK \cite{pointnetlk} pioneers the recent learning-based registration methods. It adapts PointNet \cite{pointnet} and the Lucas \& Kanade \cite{lk} algorithm into a single trainable recurrent deep neural network. Deep Closest Point (DCP) \cite{dcp} proposed to use a transformer network based on DGCNN \cite{dgcnn} to extract features, and train the network end-to-end by back-propagating through the SVD layer. PRNet \cite{prnet} tries to extend DCP to an iterative pipeline and deals with partially overlapping point cloud registration.

\subsubsection{Learning on Point Cloud}
\label{sec:pointcloud-related}
Recently a large volume of research papers apply deep learning techniques for learning on point clouds. Volumetric methods \cite{voxnet,voxelnet} apply discrete 3D convolution on the voxel representation. OctNet \cite{octnet} and O-CNN \cite{ocnn} try to design efficient high-resolution 3D convolution using the sparsity property of point clouds. Other methods \cite{pointconv,pointcnn,kpconv,interpconv} try to directly define convolution in the continuous Euclidean space, or convert the point clouds to a new space for implementing easy convolution-like operations \cite{spectralcnn,splatnet}. Contrary to the effort of adapting convolution to point clouds, PointNet \cite{pointnet} and PointNet++ \cite{pointnet++}, which use simple permutation invariant pooling operations to aggregate information from individual points, are widely used recently due to their simplicity. \cite{dgcnn,rscnn} view point clouds as graphs with neighbors connecting to each other, and apply graph neural networks (GNN) \cite{gnn} to extract features.

\begin{figure}
\centering
\includegraphics[width=0.95\linewidth]{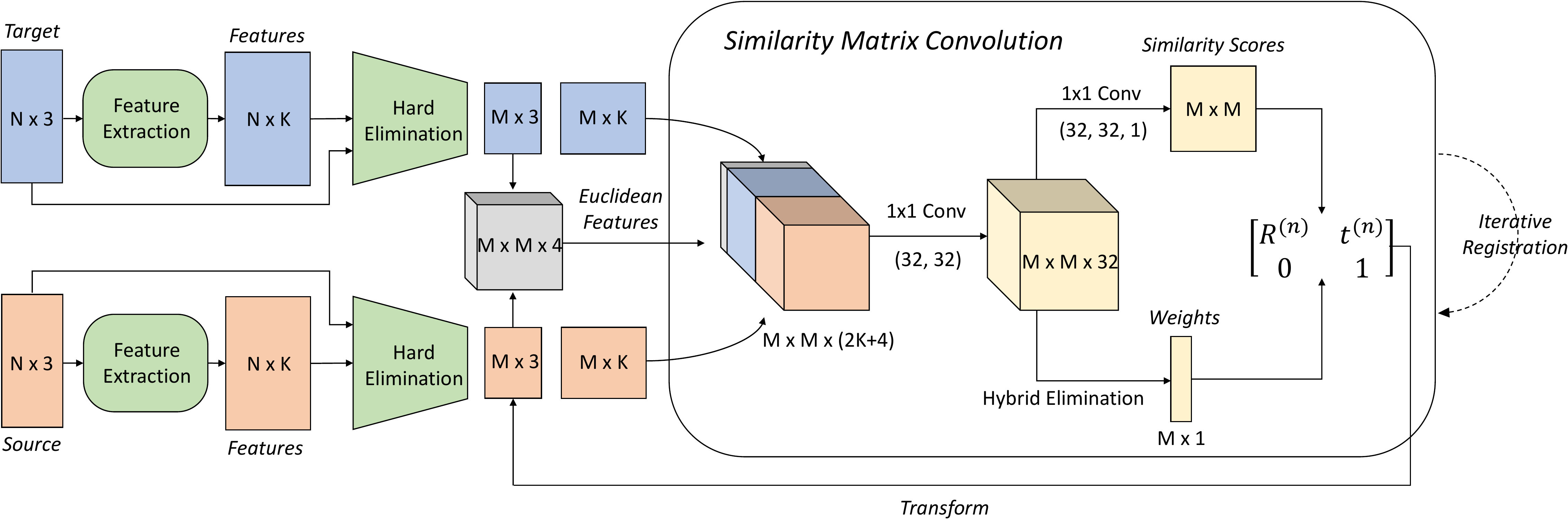}
\caption{The overall architecture of the IDAM registration pipeline. Details of hard point elimination and hybrid point elimination are demonstrated in \reffig{elimination}.}
\label{fig:model}
\end{figure}

\section{Model}
\label{sec:model}

This section describes the proposed IDAM point cloud registration model. The diagram of the whole pipeline is shown in \reffig{model}. The details of each component is explained in the following sections.

\subsection{Notation}
\label{sec:notation}

\newcommand{\source}{\mathcal{S}}
\newcommand{\target}{\mathcal{T}}
\newcommand{\numSourcePoint}{N_\source}
\newcommand{\numTargetPoint}{N_\target}
\newcommand{\pointInSource}{\mathbf{p}}
\newcommand{\pointInTarget}{\mathbf{q}}
\newcommand{\trueR}{\mathbf{R}^*}
\newcommand{\truet}{\mathbf{t}^*}
\newcommand{\rotation}{\mathbf{R}}
\newcommand{\translation}{\mathbf{t}}

Here we introduce some notation that will be used throughout the paper. The problem of interest is that for a given source point cloud $\source$ of $\numSourcePoint$ points and a target point cloud $\target$ of $\numTargetPoint$ points, we need to find the ground truth rigid body transformation $(\trueR, \truet)$ that aligns $\source$ to $\target$. Let $\pointInSource_i\in\source$ denote the $i$th point in the source, and $\pointInTarget_j\in\target$ the $j$th point in the target.

\subsection{Similarity Matrix Convolution}
\label{sec:smc}

To find the rigid body transformation $\trueR$ and $\truet$, we need to find  a set of point correspondences between the source and target point clouds. Most of the existing methods achieve this by using the inner product (or $L2$ distance) of the point features as a measure of similarity, and directly pick the ones with the highest (or lowest for $L2$) response. 

However, this has two shortcomings. First of all, one point in $\source$ may have multiple possible correspondences in $\target$, and one-shot matching is not ideal since the points chosen as correspondences may not be the correct ones due to randomness. Inspired by ICP, we argue that incorporating distance information between points into an iterative matching processing can alleviate this problem, since after an initial registration,  correct point correspondences are more likely to be closer to each other. 

The second drawback of direct feature similarity computation is that it has limited power of identifying the similarity between two points because the way of matching is the same for different pairs. Instead, we have a learned network that accepts the whole feature vectors and outputs the similarity scores. This way, the network takes into consideration the combinations of features from two points in a pair for matching.

\newcommand{\featureForSource}{\mathbf{u}^\source}
\newcommand{\featureForTarget}{\mathbf{u}^\target}
\newcommand{\featureMatrix}{\mathbf{T}^{(n)}}
\newcommand{\similarityMatrix}{\mathbf{S}^{(n)}}
\newcommand{\Ratn}{\mathbf{R}^{(n)}}
\newcommand{\tatn}{\mathbf{t}^{(n)}}

Based on the above intuition, we propose \textbf{distance-aware similarity matrix convolution} for finding point correspondences. Suppose we have the geometric features $\featureForSource(i)$ for $\pointInSource_i\in\source$ and $\featureForTarget(j)$ for $\pointInTarget_j\in\target$, both with dimension $K$. We form the \textbf{distance-augmented feature tensor} at iteration $n$ as

\begin{align}
  \featureMatrix(i,j) = [\featureForSource(i); \featureForTarget(j); \norm{\pointInSource_i-\pointInTarget_j}; \frac{\pointInSource_i-\pointInTarget_j}{\norm{\pointInSource_i-\pointInTarget_j}}]
\end{align}
where $[\cdot;\cdot]$ denotes concatenation. The $(2K+4)$-dimensional vector at the $(i,j)$ location of $\featureMatrix$ is a combination of the geometric and Euclidean features for the point pair $(\pointInSource_i,\pointInTarget_j)$. The $4$-dimensional Euclidean features comprise the distance between $\pointInSource_i$ and $\pointInTarget_j$, and the unit vector pointing from $\pointInTarget_j$ to $\pointInSource_i$. Each augmented feature vector in $\featureMatrix$ encodes the joint information of the local shapes of the two points and their current relative position, which are useful for computing similarity scores at each iteration.

The distance-augmented feature tensor $\featureMatrix$ can be seen as a $(2K+4)$-channel 2D image. To extract a similarity score for each point pair, we apply a series of $1\times 1$ 2D convolution on $\featureMatrix$ that outputs a single channel image of the same spatial size at the last layer. This is equivalent to applying a multi-layer perceptron on the augmented feature vector at each position. Then we apply a Softmax function on each row of the single channel image to get the \textbf{similarity matrix}, denoted as $\similarityMatrix$. $\similarityMatrix(i,j)$ represents the ``similarity score'' (the higher the more similar) for $\pointInSource_i$ and $\pointInTarget_j$. Each row of $\similarityMatrix$ defines a normalized probability distribution over all the points in $\target$ for some $\pointInSource\in\source$. As a result, $\similarityMatrix(i,j)$ can also be interpreted as the probability of $\pointInTarget_j$ being the correspondence of $\pointInSource_i$. The $1\times 1$ convolutions learn their weights using the \textbf{point matching loss} described in \refsec{matching-loss}. They learn to take into account the interaction between the shape and distance information to output a more accurate similarity score compared to simple inner product.

To find the correspondence pairs, we take the argmax of each row of $\similarityMatrix$. The results are a set of correspondence pairs $\set{(\pointInSource_i, \pointInSource^\prime_i)}{\forall \pointInSource_i\in\source}$, with which we solve the following optimization problem to find the estimated rigid transformation $(\Ratn, \tatn)$

\begin{align}
\label{eq:objective-no-weight}
  \Ratn, \tatn = \argmin_{\rotation, \translation} \sum_i  \norm{\rotation\pointInSource_i+\translation-\pointInSource^\prime_i}^2
\end{align}

This is a classical absolute orientation problem \cite{horn87}, which can be efficiently solved with the orthogonal Procrustes algorithm \cite{golubmatrix,svd} using Singular Value Decomposition (SVD). $\Ratn$ and  $\tatn$ are then used to transform the source point cloud to a new position before entering the next iteration. The final estimate for $(\trueR, \truet)$ is the composition of the intermediate $(\Ratn, \tatn)$ for all the iterations.

\begin{figure}
\centering
\includegraphics[width=0.99\linewidth]{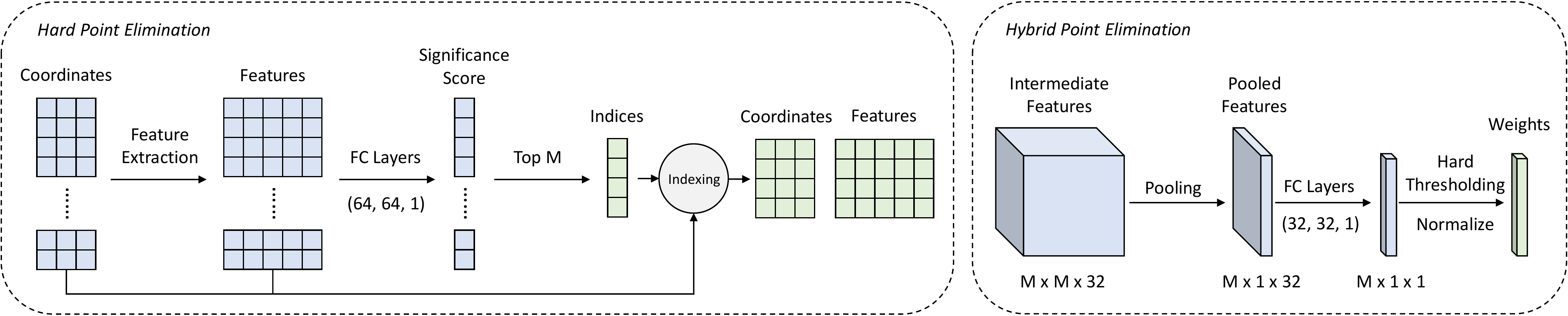}
\caption{Comparison of hard point elimination and hybrid point elimination. Hard point elimination filters points based on the features extracted independently for each point, while hybrid point elimination utilizes the joint information of the point pairs to compute weights for the orthogonal Procrustes algorithm.}
\label{fig:elimination}
\end{figure}

\subsection{Two-stage Point Elimination}
\label{sec:elimination}

Although similarity matrix convolution is powerful in terms of matching, it is computationally expensive to apply convolution on the large $\numSourcePoint\times\numTargetPoint\times(2K+4)$ tensor, because $\numSourcePoint$ and $\numTargetPoint$ are typically more than a thousand. However, if we randomly down-sample the point clouds, the performance of the model would degrade drastically since many points no longer have correspondences. To tackle this dilemma, we propose a \textbf{two-stage point elimination} process. It consists of \textbf{hard point elimination} and \textbf{hybrid point elimination} (\reffig{elimination}), which targets on improving efficiency and accuracy respectively. While manually labelling keypoints for point clouds is not practical, we propose a \textbf{mutual-supervision loss}, that uses the information in the similarity matrices $\similarityMatrix$ to supervise the point elimination modules. The details of the mutual-supervision loss is described in \refsec{loss}. In this section, we present the point elimination process for inference.

\subsubsection{Hard Point Elimination}
\label{sec:hard-elimination}

\newcommand{\sourceSampled}{\mathcal{B}_\mathcal{S}}
\newcommand{\targetSampled}{\mathcal{B}_\mathcal{T}}

To reduce the computational burden of similarity matrix convolution, we first propose the \textbf{hard point elimination} (\reffig{elimination} Left). Given the extracted local shape features for each point, we apply a multi-layer perceptron on the feature vector, and output a \textbf{significance score}. A high score means a more prominent point, such as a corner point, that can be matched with high confidence later (see the Appendix for visualization). It filters out those points in the ``flat'' regions that are ambiguous during matching. This process is done on individual points, and does not take into account the point pair information as in similarity matrix convolution. As a result, it is efficient to compute the significance score. We preserve the $M$ points for each point cloud with highest significance scores, and eliminate the remaining points. In our network, we choose $M=\ceil{\frac{N}{6}}$, where $N$ can be $\numSourcePoint$ or $\numTargetPoint$. Denote the set of points in $\source$ preserved by hard point elimination as $\sourceSampled$, and that for the target as $\targetSampled$.

\subsubsection{Hybrid Point Elimination}
\label{sec:hybrid-elimination}

\newcommand{\intermediateFeatureMatrix}{\mathbf{F}}
While hard point elimination improves the efficiency significantly, it has negative effect on the performance of the model. The correct corresponding point in the target point cloud for a point in the source point cloud may be mistakenly eliminated in hard elimination. Therefore, similarity matrix convolution will never be able to find the correct correspondence. However, since we always try to find the correspondence with the maximal similarity score for every point in the source, these ``negative'' point pairs can make the rigid body transformation obtained by solving \refeq{objective-no-weight} inaccurate. This problem is especially severe when the model is dealing with two point clouds that only partially overlap with each other. In this case, even without any elimination, some points will not have any correspondence whatsoever.

To alleviate this problem, we propose a \textbf{hybrid point elimination} (\reffig{elimination} Right) process, applied after similarity matrix convolution. Hybrid point elimination is a mixture of both hard elimination and soft elimination, and operates on point pairs instead of individual points. It uses a permutation-invariant pooling operation to aggregate information across all possible correspondences for a given point in the source, and outputs the \textbf{validity score}, for which a higher score means higher probability of having a true correspondence.  Formally, let $\intermediateFeatureMatrix$ be the intermediate output (see \reffig{model}) of the similarity matrix convolution of shape $M\times M\times K^\prime$. Hybrid point elimination first computes the validity score

\begin{align}
v(i) = \sigma(f(\bigoplus_{j}(\intermediateFeatureMatrix(i,j))))
\end{align}
where $\sigma(\cdot)$ is the sigmoid function, $\bigoplus$ is an element-wise permutation invariant pooling method, such as ``mean'' or ``max'', and $f$ is a multi-layer perceptron that takes the pooled features as input and outputs the scores. This permutation invariant pooling technique is used in a variety point cloud processing \cite{pointnet,pointnet++} and graph neural network \cite{dgcnn,rscnn} models. Following \cite{pointnet,pointnet++} we use element-wise max for $\bigoplus$. This way, we have a validity score for each point in the source, and thus for each point pair. It can be seen as the probability that a correspondence pair is correct.

With this validity score, we then compute the \textbf{hybrid elimination weights}. The weight for the $i$th point pair is defined as

\begin{align}
w_i = \frac{v(i) \cdot \indicator{v(i) \geq \text{median}_{k}(v(k))}}{\sum_i v(i) \cdot \indicator{v(i) \geq \text{median}_{k}(v(k))}}
\end{align}
where $\indicator{\cdot}$ is the indicator function. What this weighting process does is that it gives 0 weight to those points with lowest validity scores (hard elimination), and weighs the rest proportionally to the validity scores (soft elimination). With this elimination weight vector, we can obtain the $(\Ratn, \tatn)$ with a slightly different objective function from \refeq{objective-no-weight}

\begin{align}
\label{eq:objective}
  \Ratn, \tatn = \argmin_{\rotation, \translation} \sum_i w_i \norm{\rotation\pointInSource_i+\translation-\pointInSource^\prime_i}^2
\end{align}

This can still be solved using SVD with little overhead. Ideally, the hybrid point elimination can eliminate those point pairs that are not correct due to noise and incompletion, giving better performance on estimating $\Ratn$ and $\tatn$ (see the Appendix for visualization).

\subsection{Mutual-Supervision Loss}
\label{sec:loss}

In this section, we describe in detail the \textbf{mutual-supervision loss} that is used to train the network. With this loss, we can train the similarity matrix convolution, along with the two-stage elimination module, without extra annotations of keypoints. The loss is the sum of three parts, which will be explained in the following.

Note that training on all the points during each forward-backward loop is inefficient and unnecessary. However, since hard point elimination does not function properly during training yet, we do not have direct access to $\sourceSampled$ and $\targetSampled$ (see the definitions in \refsec{hard-elimination}). Therefore, we need some way to sample points from the source and the target for training. This sampling technique is described in \refsec{sampling}. In this section we accept that as given, and abuse the notation $\sourceSampled$ for the \textbf{source sampled set} and $\targetSampled$ for the \textbf{target sampled set}, which both contain the $M$ sampled points for training. Let $\pointInSource_i$ denote the $i$th point in $\sourceSampled$ and $\pointInTarget_j$ denote the $j$th point in $\targetSampled$

\subsubsection{Point Matching Loss}
\label{sec:matching-loss}

The point matching loss is used to supervise the similarity matrix convolution. It is a standard cross-entropy loss. The point matching loss for the $n$th iteration is defined as

\begin{align}
\mathcal{L}^{(n)}_{\text{match}}(\source, \target, \trueR, \truet)=\frac{1}{M}\sum_{i=1}^{M}-\log(\similarityMatrix(i, j^*))\cdot\indicator{\norm{\trueR \pointInSource_i+\truet-\pointInTarget_{j^*}}^2\leq r^2}
\end{align}
where
\begin{align}
j^*=\argmin_{1\leq j\leq M} \norm{\trueR \pointInSource_i+\truet-\pointInTarget_j}^2
\end{align}
is the index of the point in the target sampled set $\targetSampled$ that is closest to the $i$th point in the source sampled set $\sourceSampled$ under the ground truth transformation. $r$ is hyper-parameter that controls the minimal radius within which two points are considered close enough. If the distance of $\pointInSource_i$ and $\pointInTarget_{j^*}$ is larger than $r$, they can not be seen as correspondences, and no supervision signal is applied on them. This happens frequently when the model is dealing with partially overlapping point clouds. The total point matching loss is the average of those for all the iterations.

\subsubsection{Negative Entropy Loss}
\label{sec:hard-elimination-loss}

This loss is used for training hard point elimination. The problem for training hard point elimination is that we do not have direct access to annotations of keypoints. Therefore, we propose to use a \textbf{mutual supervision} technique, which uses the result of the point matching loss to supervise hard point elimination. This mutual supervision is based on the intuition that if a point $\pointInSource_i\in\sourceSampled$ is a prominent point (high significance score), the probability distribution defined by the $i$th row of $\similarityMatrix$ should have low entropy because it is confident in matching. On the other hand, the supervision on the similarity matrices has no direct relationship to hard point elimination. Therefore, the \textbf{negative entropy} of the probability distribution can be seen as a supervision signal for the significance scores. Mathematically, the \textbf{negative entropy loss} for the $n$th iteration can be defined as

\begin{align}
\mathcal{L}^{(n)}_{\text{hard}}(\source, \target, \trueR, \truet)=\frac{1}{M}\sum_{i=1}^{M}|s(i)-\sum_{j=1}^{M}\similarityMatrix(i, j)\log(\similarityMatrix(i, j))|^2
\end{align}
where $s(i)$ is the significant score for the $i$th point in $\sourceSampled$. Although this loss can be defined for any iteration, we only use the one for first iteration, because in the early stages of registration the shape features are more important than the Euclidean features. We want the hard point elimination module learns to filter points only based on shape information. We cut the gradient flow from the negative entropy loss to $\similarityMatrix$ to prevent interference with the training of similarity matrix convolution.

\subsubsection{Hybrid Elimination Loss}
\label{sec:hybrid-elimination-loss}

A similar mutual supervision idea can also be used for training the hybrid point elimination. The difference is that hybrid elimination takes into account the point pair information, while hard point elimination only looks at individual points. As a result, the mutual supervision signal is much more obvious for hybrid point elimination. We simply use the probability that there exists a point in $\targetSampled$ which is the correspondence of point $\pointInSource_i\in\sourceSampled$ as the supervision signal for $v_i$ (validity score). Instead of computing the probability explicitly, the \textbf{hybrid elimination loss} for the $n$th iteration is defined as

\begin{align}
\mathcal{L}^{(n)}_{\text{hybrid}}(\source, \target, \trueR, \truet)=\frac{1}{M}\sum_{i=1}^{M}-\mathbbm{I}_i\cdot \log( v_i)-(1-\mathbbm{I}_i)\cdot \log (1-v_i)
\end{align}
where
\begin{align}
\mathbbm{I}_i=\indicator{\norm{\trueR\pointInSource_i+\truet-\pointInTarget_{\argmax_j \similarityMatrix(i,j)}}^2\leq r^2}
\end{align}
In effect, this loss assigns a positive label $1$ to those points in $\sourceSampled$ that correctly finds its correspondence, and a negative label $0$ to those that do not. In the long run, those point pairs with high probability of correct matching will have higher validity scores.

\subsection{Balanced Sampling for Training}
\label{sec:sampling}

In this section, we describe a balanced sampling technique to sample points for training our network. We first sample $\ceil{\frac{M}{2}}$ points from $\source$ with the following unnormalized probability distribution

\begin{align}
p_{\text{pos}}(i)=\indicator{(\min_{\pointInTarget\in\target} \norm{\trueR\pointInSource_i+\truet-\pointInTarget}^2)\leq r^2} + \epsilon
\end{align}
where $\epsilon=10^{-6}$ is some small number. This sampling process aims to randomly sample ``positive'' points from $\source$, in the sense that they indeed have correspondences in the target. It introduces the $\epsilon$ to avoid errors when encountering the singularity cases where no points in the source have correspondences in the target.

Similarly, we sample $(M-\ceil{\frac{M}{2}})$ ``negative'' points from $\source$ using the unnormalized distribution

\begin{align}
p_{\text{neg}}(i)=\indicator{(\min_{\pointInTarget\in\target} \norm{\trueR\pointInSource_i+\truet-\pointInTarget}^2)> r^2} + \epsilon
\end{align}

This way, we have a set $\sourceSampled$ of points of size $M$, with both positive and negative instances. To sample points from the target, we simply find the closest points of each point from $\sourceSampled$ in the target

\begin{align}
\targetSampled = \set{\argmin_{\pointInTarget}\norm{\trueR\pointInSource_i+\truet-\pointInTarget}}{i\in\sourceSampled}
\end{align}

This balanced sampling technique randomly samples points from $\source$ and $\target$, while keeping a balance between points that have correspondences and points that do not.

\section{Experiments}
\label{sec:experiments}

This section shows the experimental results to demonstrate the performance and efficiency of our method. We also conduct ablation study to show the effectiveness of each component of our model.

\subsection{Experimental Setup}
\label{sec:experimental-setup}

We train our model with the Adam \cite{adam} optimizer for 40 epochs. The initial learning rate is $1\times 10^{-4}$, and is multiplied by 0.1 after 30 epochs. We use a weight decay of $1\times 10^{-3}$ and no Dropout \cite{dropout}. We use the FPFH implementation from the Open3D \cite{open3d} library and a very simple graph neural network (GNN) for feature extraction. The details of the GNN architecture are described in the supplementary material. For both FPFH and GNN features, the number of iterations is set to 3.

Following \cite{prnet}, all the experiments are done on the ModelNet40 \cite{modelnet} dataset. ModelNet40 includes 9843 training shapes and 2468 testing shapes from 40 object categories. For a given shape, we randomly sample 1024 points to form a point cloud. For each point cloud, we randomly generate rotations within $[0^\circ, 45^\circ]$ and translation in $[-0.5, 0.5]$. The original point cloud is used as the source, and the transformed point cloud as the target. To generate partially overlap point clouds, we follow the same method as \cite{prnet}, which fixes a random point far away from the two point clouds, and preserve 768 points closest to the far point for each point cloud. 

We compare our method to ICP, Go-ICP, FGR, FPFH+RANSAC, PointNetLK, DCP and PRNet. All the data-driven methods are trained on the same training set. We use the same metrics as \cite{prnet,dcp} to evaluate all these methods. For the rotation matrix, the root mean square error (RMSE($\rotation$)) and mean absolute error (MAE($\rotation$)) in degrees are used. For the translation vector, the root mean square error (RMSE($\translation$)) and mean absolute error (MAE($\translation$)) are used.

\subsection{Results}
\label{sec:results}

In this section, we show the results for three different experiments to demonstrate the effectiveness and robustness of our method. These experimental settings are the same as those in \cite{prnet}. We also include in the supplementary material some visualization results for these experiments.

\subsubsection{Unseen Shapes}
\label{sec:unseen-shapes}

First, we train our model on the training set of ModelNet40 and evaluate on the test set. Both the training set and test set of ModelNet40 contain point clouds from all the 40 categories. This experiment evaluates the ability to generalize to unseen point clouds. \reftab{unseen-shapes} shows the results.

\begin{table}
\begin{center}
\caption{Results for testing on point clouds of unseen shapes in ModelNet40.}
\label{tab:unseen-shapes}
\begin{tabular}{lcccc}
\hline\noalign{\smallskip}
Model & RMSE($\rotation$) & MAE($\rotation$) & RMSE($\translation$) & MAE($\translation$) \\
\noalign{\smallskip}
\hline
\noalign{\smallskip}
ICP & 33.68 & 25.05 & 0.29 & 0.25\\
FPFH+RANSAC & 2.33 & 1.96 & \textbf{0.015} & 0.013\\
FGR & 11.24 & 2.83 & 0.030 & 0.008\\
Go-ICP & 14.0 & 3.17 & 0.033 & 0.012\\
PointNetLK & 16.74 & 7.55 & 0.045 & 0.025\\
DCP & 6.71 & 4.45 & 0.027 & 0.020\\
PRNet & 3.20 & 1.45 & 0.016 & 0.010\\
\hline
FPFH+IDAM & \textbf{2.46} & \textbf{0.56} & 0.016 & \textbf{0.003}\\
GNN+IDAM & 2.95 & 0.76 & 0.021 & 0.005\\
\hline
\end{tabular}
\end{center}
\end{table}

We can see that local registration method ICP performs poorly because the initial rotation angles are large. FPFH+RANSAC is the best performing traditional method, which is comparable to many learning-based methods. Note that both RANSAC and FGR use FPFH methods, and our method with FPFH features outperforms both of them. Neural network models have a good balance between performance and efficiency. Our method outperforms all the other methods with both hand-crafted (FPFH) and learned (GNN) features.

Surprisingly, FPFH+IDAM has better performance than GNN+IDAM. One possibility is that the GNN overfits to the point clouds in the training set, and does not generalize well to unseen shapes. However, as will be shown in later sections, GNN+IDAM is more robust to noise and also more efficient that FPFH+IDAM.

\subsubsection{Unseen Categories}
\label{sec:unseen-categories}

In the second experiment, we use the first 20 categories in the training set of ModelNet40 for training, and evaluate on the other 20 categories on the test set. This experiment tests the capability to generalize to point clouds of unseen categories. The results are summarized in \reftab{unseen-categories}. 
We can see that without training on the testing categories, all the learning-based methods perform worse consistently. Traditional methods are not affected that much as expected. Based on different evaluation metrics, FPFH+RANSAC and FPFH+IDAM are the best performing methods.
\begin{table}
\begin{center}
\caption{Results for testing on point clouds of unseen categories in ModelNet40. }
\label{tab:unseen-categories}
\begin{tabular}{lcccc}
\hline\noalign{\smallskip}
Model & RMSE($\rotation$) & MAE($\rotation$) & RMSE($\translation$) & MAE($\translation$) \\
\noalign{\smallskip}
\hline
\noalign{\smallskip}
ICP & 34.89 & 25.46 & 0.29 & 0.25\\
FPFH+RANSAC & \textbf{2.11} & 1.82 & \textbf{0.015} & 0.013\\
FGR & 9.93 & 1.95 & 0.038 & 0.007\\
Go-ICP & 12.53 & 2.94 & 0.031 & 0.010\\
PointNetLK & 22.94 & 9.66 & 0.061 & 0.033\\
DCP & 9.77 & 6.95 & 0.034 & 0.025\\
PRNet & 4.99 & 2.33 & 0.021 & 0.015\\
\hline
FPFH+IDAM & 3.04 & \textbf{0.61} & 0.019 & \textbf{0.004}\\
GNN+IDAM & 3.42 & 0.93 & 0.022 & 0.005\\
\hline
\end{tabular}
\end{center}
\end{table}

\subsubsection{Gaussian Noise}
\label{sec:gaussian-noise}

In the last experiment, we add random Gaussian noise with standard deviation 0.01 to all the shapes, and repeat the first experiment (unseen shapes). The random noise is clipped to $[-0.05, 0.05]$. As shown in \reftab{gaussian-noise}, both traditional methods and IDAM based on FPFH features perform much worse than the noise-free case. This demonstrates that FPFH is not very robust to noise. The performance of data-driven methods are comparable to the noise-free case, thanks to the powerful feature extraction networks. Our method based on GNN features has the best performance compared to others.
\begin{table}
\begin{center}
\caption{Results for testing on point clouds of unseen shapes in ModelNet40 with Gaussian noise. }
\label{tab:gaussian-noise}
\begin{tabular}{lcccc}
\hline\noalign{\smallskip}
Model & RMSE($\rotation$) & MAE($\rotation$) & RMSE($\translation$) & MAE($\translation$) \\
\noalign{\smallskip}
\hline
\noalign{\smallskip}
ICP & 35.07 & 25.56 & 0.29 & 0.25\\
FPFH+RANSAC & 5.06 & 4.19 & 0.021 & 0.018\\
FGR & 27.67 & 13.79 & 0.070 & 0.039\\
Go-ICP & 12.26 & 2.85 & 0.028 & 0.029\\
PointNetLK & 19.94 & 9.08 & 0.057 & 0.032\\
DCP & 6.88 & 4.53 & 0.028 & 0.021\\
PRNet & 4.32 & 2.05 & \textbf{0.017} & 0.012\\
\hline
FPFH+IDAM & 14.21 & 7.52 & 0.067 & 0.042\\
GNN+IDAM & \textbf{3.72} & \textbf{1.85} & 0.023 & \textbf{0.011}\\
\hline
\end{tabular}
\end{center}
\end{table}

\subsection{Efficiency}
\label{sec:efficiency}

We test the speed of our method, and compare it to ICP, FGR, FPFH+RANSAC, PointNetLK, DCP and PRNet. We use the Open3D implementation of ICP, FGR and FPFH+RANSAC, and the official implementation of PointNetLK, DCP and PRNet released by the authors. The experiments are done on a machine with 2 Intel Xeon Gold 6130 CPUs and a single Nvidia GeForce RTX 2080 Ti GPU. We use a batch size of 1 for all the neural network based models. The speed is measured in seconds per frame.

We test the speed on point clouds with 1024, 2048 and 4096 points, and the results are summarized in \reftab{efficiency}. It can be seen that neural network based methods are generally faster than traditional methods. When the number of points is small, IDAM with GNN features is only slower than DCP. But as the number of points increases, IDAM+GNN is much faster than all the other methods. Although FPFH+RANSAC has the best performance among non-learning methods, it is also the slowest. Note that our method with FPFH features is $2\times$ to $5\times$ faster than the other two methods (FGR and RANSAC) which also use FPFH.

\begin{table}
\begin{center}
\caption{Comparison of speed of different models. IDAM(G) and IDAM(F) represent GNN+IDAM and FPFH+IDAM respectively. RANSAC also uses FPFH. Speed is measured in seconds-per-frame.}
\label{tab:efficiency}
\begin{tabular}{lcccccccc}
\hline\noalign{\smallskip}
 & IDAM(G) & IDAM(F) & ICP & FGR & RANSAC & PointNetLK & DCP & PRNet \\
\noalign{\smallskip}
\hline
\noalign{\smallskip}
1024 points & 0.026 & 0.050 & 0.095 & 0.123 & 0.159 & 0.082 & 0.015 & 0.022\\
2048 points & 0.038 & 0.078 & 0.185 & 0.214 & 0.325 & 0.085 & 0.030 & 0.048\\
4096 points & 0.041 & 0.175 & 0.368 & 0.444 & 0.685 & 0.098 & 0.084 & 0.312\\
\hline
\end{tabular}
\end{center}
\end{table}

\subsection{Ablation Study}
\label{sec:ablation-study}

In this section, we present the results of ablation study of IDAM to show the effectiveness of each component. We examine three key components of our model: distance-aware similarity matrix convolution (denoted as SM), hard point elimination (HA) and hybrid point elimination (HB). We use BS to denote the model that does not contain any of the three components mentioned above. Since hard point elimination is necessary for similarity matrix convolution due to memory constraints, we replace it with random point elimination in BS. We use inner-product of features in BS when similarity matrix convolution is disabled. As a result, BS is just a simple model that uses the inner-product of features as similarity scores to find correspondences, and directly solves the absolute orientation problem (\refeq{objective-no-weight}). We add the components one by one and compare their performance for GNN features. We conducted the experiments under the settings of ``unseen categories'' as described in \refsec{unseen-categories}. The results are summarized in \reftab{ablation}.

It can be seen that even with random sampling, similarity matrix convolution already outperforms the baseline (BS) by a large margin. The two-stage point elimination (HA and HB) further boosts the performance significantly.

\begin{table}
\begin{center}
\caption{Comparison of the performance of different model choices for IDAM. These experiments examine the effectiveness of similarity matrix convolution (SM), hard point elimination (HA) and hybrid point elimination (HB).}
\label{tab:ablation}
\begin{tabular}{lcccc}
\hline\noalign{\smallskip}
Model & RMSE($\rotation$) & MAE($\rotation$) & RMSE($\translation$) & MAE($\translation$) \\
\noalign{\smallskip}
\hline
\noalign{\smallskip}
BS & 7.77 & 5.33 & 0.055 & 0.047\\
BS+SM & 5.08 & 3.58 & 0.056 & 0.042\\
BS+HA+SM & 4.31 & 2.89 & 0.029 & 0.019\\
BS+HA+SM+HB & \textbf{3.42} & \textbf{0.93} & \textbf{0.022} & \textbf{0.005}\\
\hline
\end{tabular}
\end{center}
\end{table}

\section{Conclusions}
\label{sec:conclusion}

In this paper, we propose a novel data-driven pipeline named IDAM for partially overlapping 3D point cloud registration. We present a novel distance-aware similarity matrix convolution to augment the network's ability of finding correct correspondences in each iteration. Moreover, a novel two-stage point elimination method is proposed to improve performance while reducing computational complexity. We design a mutual-supervised loss for training IDAM end-to-end without extra annotations of keypoints. Experiments show that our method performs better than the current state-of-the-art point cloud registration methods and is robustness to noise.

\subsubsection{Acknowledgements}
\label{sec:acknowledgements}
This work was supported in part by the National Key Research and Development Program of China under Grant 2017YFA0700800.

%
%
\bibliographystyle{splncs04}
\bibliography{4625}


\section*{Supplementary material}

# Appendix A  Graph Neural Network

In this section, we describe the details of the graph neural network (GNN) we use in our IDAM+GNN model.

Suppose we have a point cloud $\mathcal{P}$. Let $\mathcal{N}_i$ denote the set of indices of $K$ points closest to the point $\mathbf{p}_i$ in $\mathcal{P}$. In the $n$th layer of GNN, let $\mathbf{u}_i^{(n)}$ be the feature vector for $\mathbf{p}_i$. Then the features for the next layer is computed as

$$\mathbf{u}_i^{(n+1)} = f(\bigoplus_{j \in \mathcal{N}_i} g(\mathbf{u}_i^{(n)} - \mathbf{u}_j^{(n)})) \tag{1}$$

where $f$ and $g$ are multi-layer perceptrons (MLP), $\bigoplus$ is the element-wise max operation. We use a 2-hidden-layer MLP with output dimensions (64, 64) for $g$, and a single-layer MLP with output dimension 64 for $f$. Batch normalization [3] and ReLU [2] are used in each layer in the MLPs.

The initial features $\mathbf{u}_i^{(0)}$ are just the 3D coordinates of the points in $\mathcal{P}$. We stack 5 layers of Eq. 1 to hierarchically extract high-level features. The dimension of the output features is 64.

# Appendix B  Visualization

In this section, we show some visualization results. Appendix B.1 shows the visualization of IDAM+GNN registering two partially overlapping point clouds in ModelNet40. Appendix B.2 shows the results of the same model on the Stanford 3D Scan dataset [1]. Appendix B.3 shows the visualization of the preserved points by hard point elimination. Appendix B.4 visualizes how the hybrid point elimination works.

### Appendix B.1  ModelNet40

In this section, We show the visualization of IDAM+GNN registering two partially overlapping point clouds in ModelNet40. The model is trained on the whole training set of ModelNet40 with Gaussian noise of standard deviation 0.01. The visualization shows the results of the model on the shapes in the test set of ModelNet40. This corresponds to the "Gaussian Noise" experiment in the paper. The results are shown in Fig. 1.

### Appendix B.2  Stanford 3D Scan

We also test the same model (Appendix B.1) on the Stanford 3D Scan dataset [1]. We use the same method described in the paper to generate partially overlapping point clouds. Note that we only train on the ModelNet40 dataset. This shows the generalization ability of our model. The results are shown in Fig. 2.

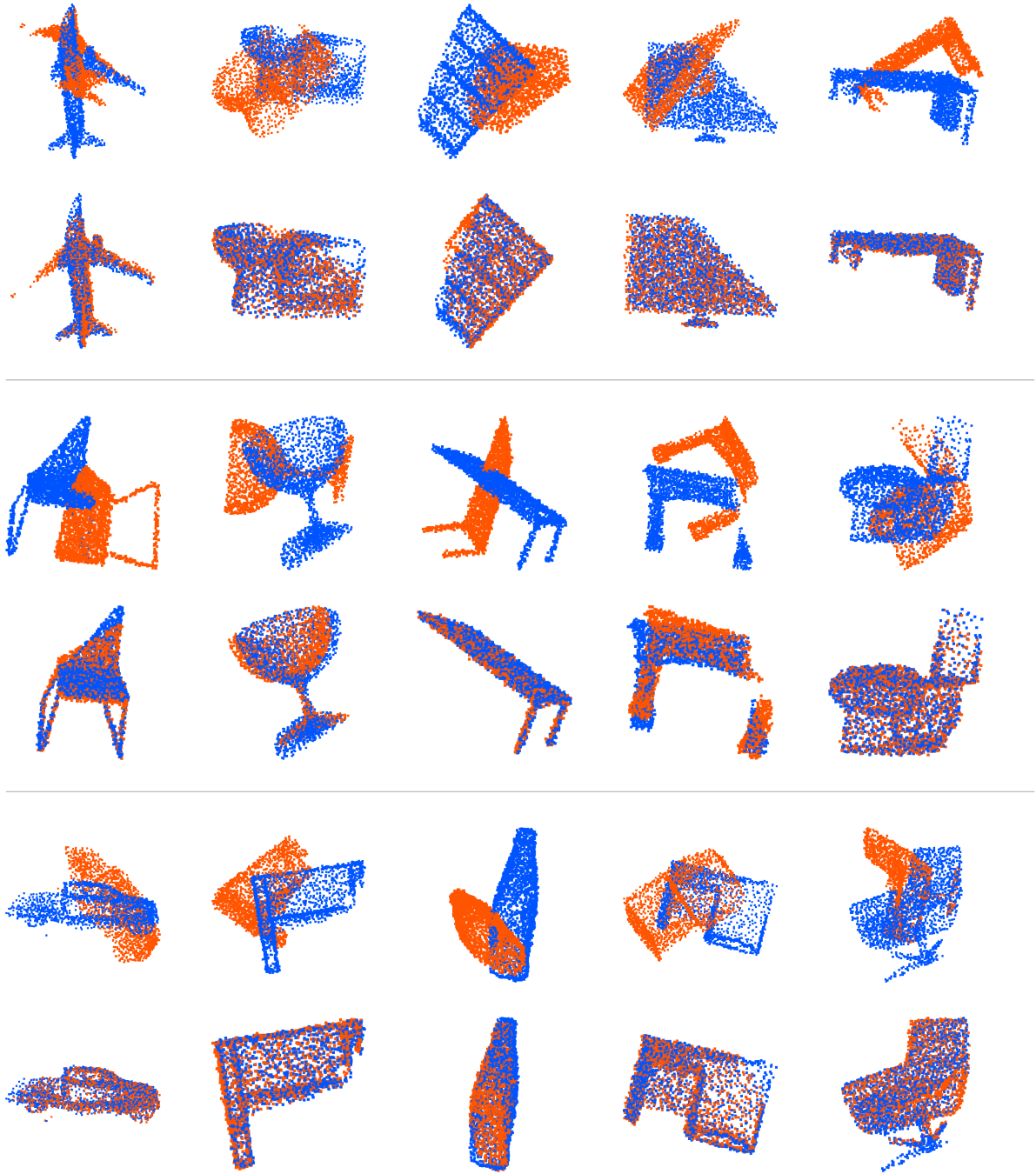

Fig. 1: The results of IDAM+GNN on ModelNet40. In each cell separated by the horizontal line, the top row shows the initial positions of the two point clouds, and the bottom row shows the results of registration.

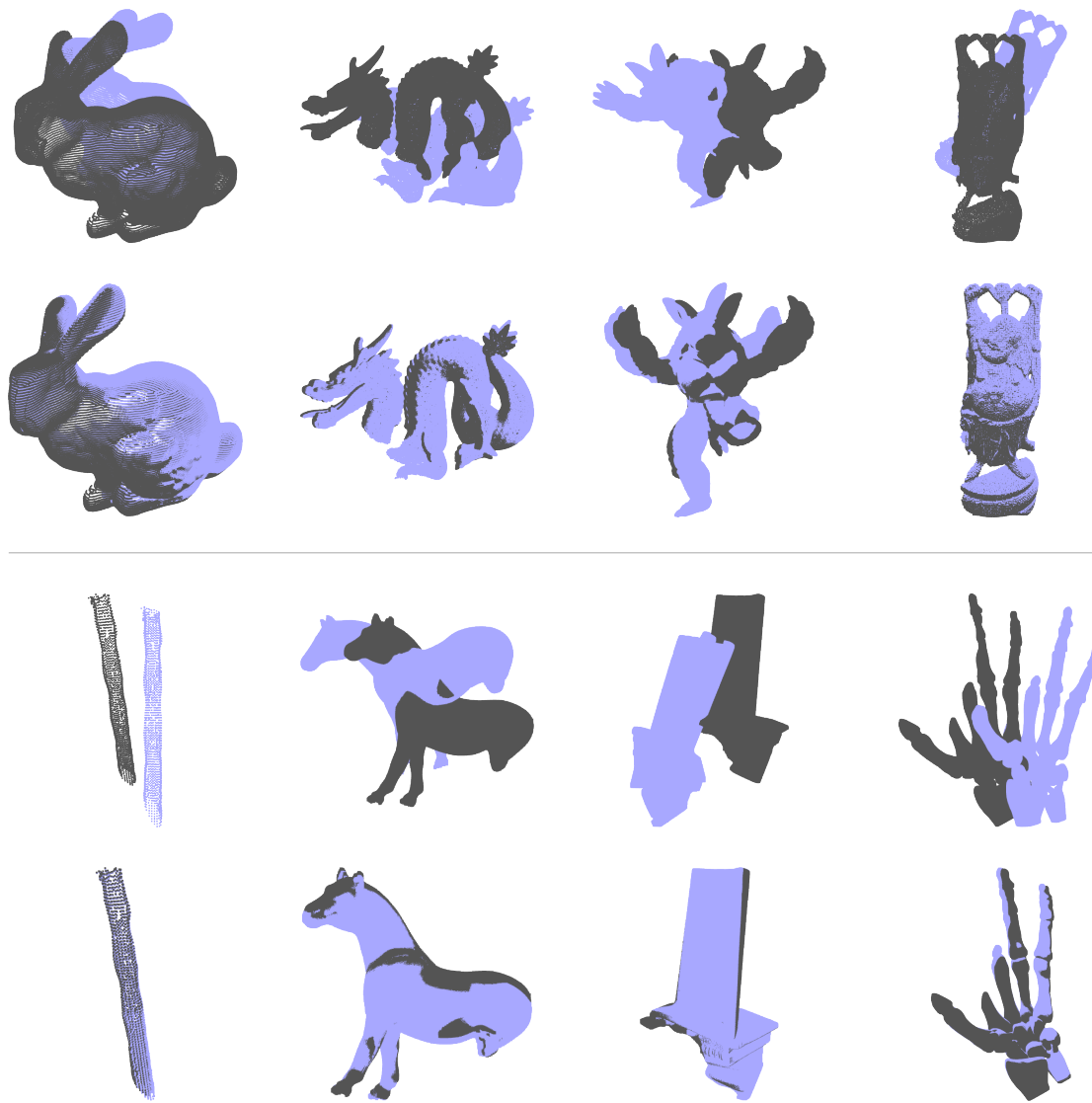

Fig. 2: The results of IDAM+GNN on the Stanford 3D Scan dataset. The model is trained on the training set on ModelNet40, and no fine-tuning is done on Stanford 3D Scan. In each cell separated by the horizontal line, the top row shows the initial positions of the two point clouds, and the bottom row shows the results of registration.

### Appendix B.3  Hard Point Elimination

This section shows the visualization results for the hard point elimination Fig. 3. In each point cloud of 1536 points, the top 128 points with the highest significance scores are highlighted.

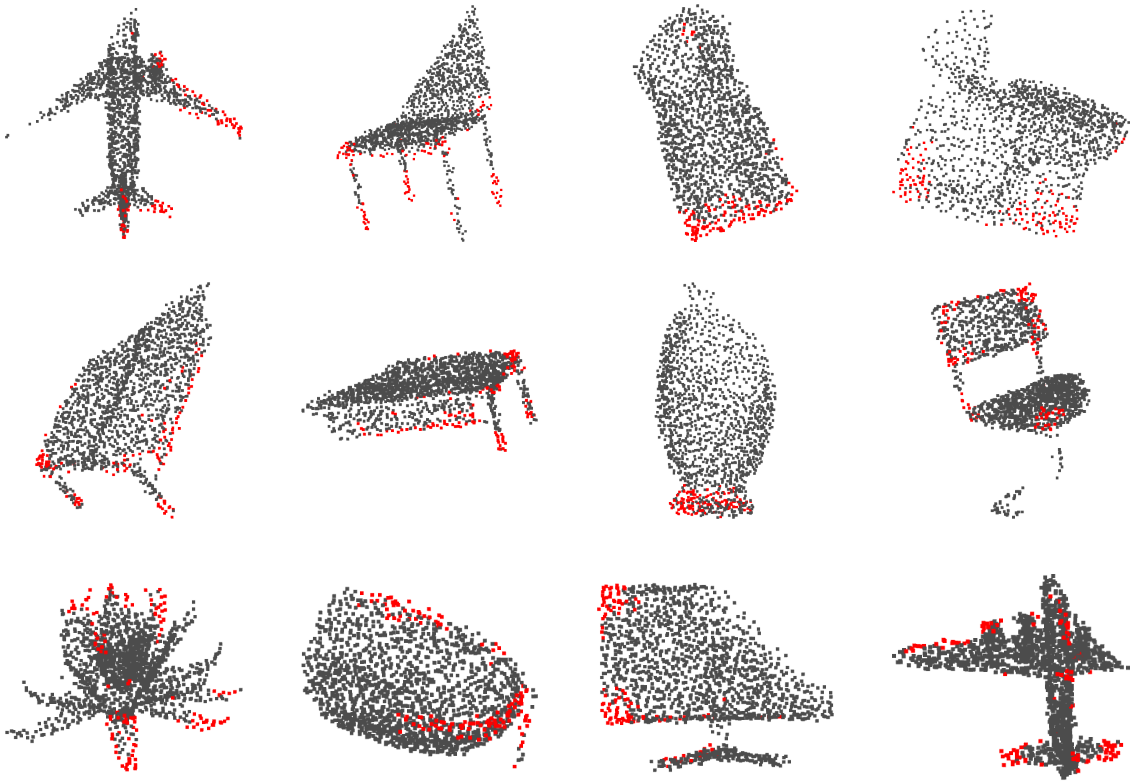

Fig. 3: Visualization of the points preserved by hard point elimination. Each point cloud contains 1536 points, and the top 128 points with the highest significance scores are highlighted.

### Appendix B.4  Hybrid Point Elimination

Fig. 4 shows the visualization of hybrid point elimination. For each point cloud pair, the correspondence pairs with non-zero weights are highlighted in green and those with zero weights are highlighted in red. We can see that most of the false positive correspondences are connected by red lines, and are thus filtered out by hybrid point elimination. This shows the effectiveness of hybrid point elimination in eliminating low-quality point pairs.

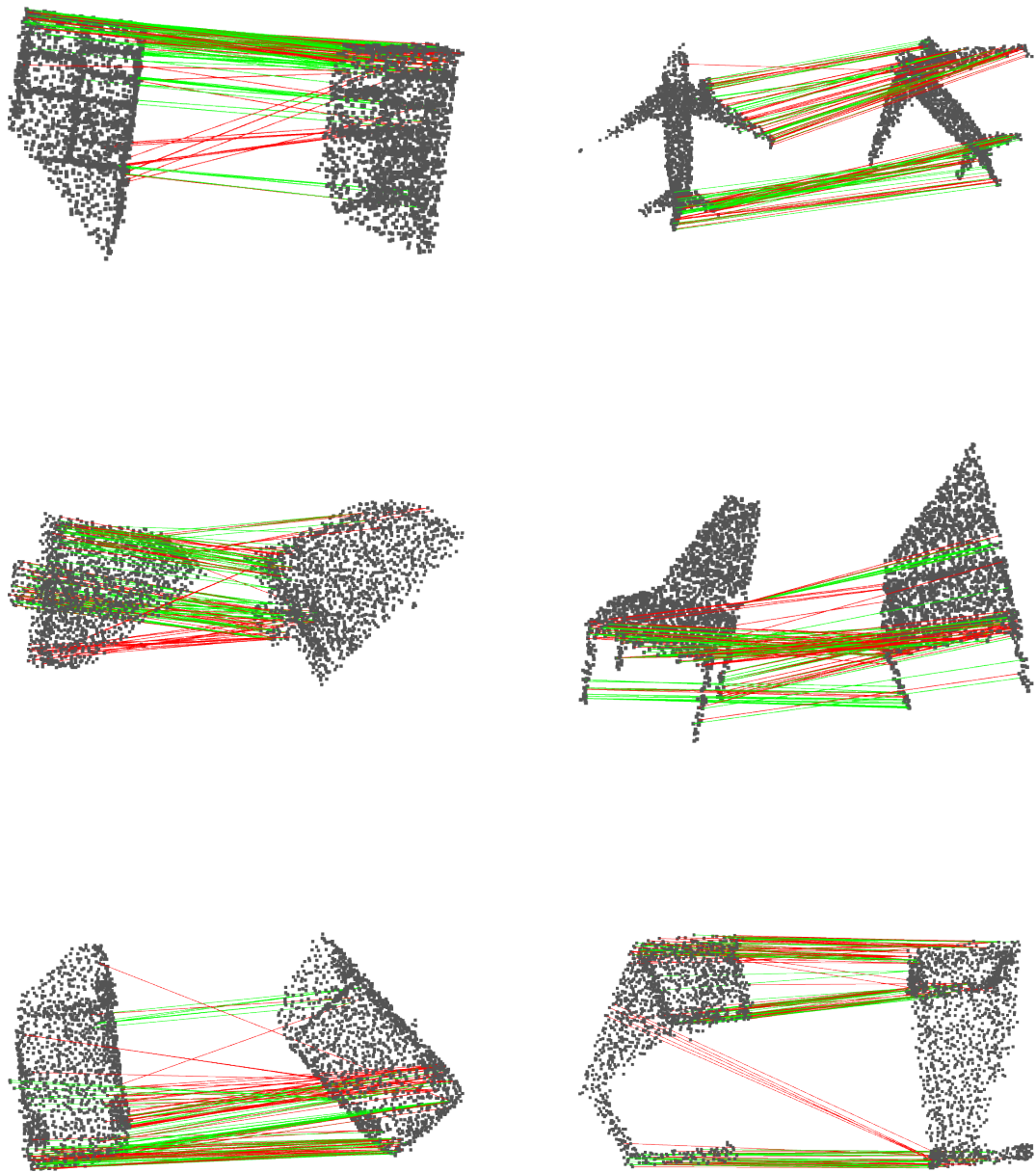

Fig. 4: Visualization of hybrid point elimination. The correspondence pairs with non-zero weights are highlighted in green and those with zero weights are highlighted in red.


\end{document}